\documentclass[runningheads]{llncs}
\usepackage[T1]{fontenc}
\usepackage{hyperref}
\usepackage{amsmath}  %
\usepackage{amssymb}
\usepackage{float}
\usepackage{hyperref}
\usepackage{graphicx}
\usepackage{amsfonts}
\usepackage[caption=false]{subfig}
\usepackage{url}
\usepackage{multirow}
\usepackage{color}
\definecolor{wb}{rgb}{1, 0, 0}

\newcommand{\stack}[1]{\!\!\begin{array}{c}\scriptstyle #1\end{array}\!\!}
\newcommand{\brs}[1][-1mm]{\\[#1]\scriptstyle}
\begin{document}
\title{Improving Robustness of AlphaZero Algorithms to Test-Time Environment Changes}

\titlerunning{Making AlphaZero Robust to Environment Changes}
\authorrunning{I. Tamassia and W. Böhmer}
\author{Isidoro Tamassia\inst{1,2,}\thanks{Work carried out while affiliated with TU Delft.} 
\and
Wendelin Böhmer\inst{1}}
\institute{Department of Intelligent Systems, TU Delft, The Netherlands \and
Department of Computer Science, KU Leuven, Belgium
\email{isidoro.tamassia@kuleuven.be}, \email{j.w.bohmer@tudelft.nl}\\
}

\maketitle              %
\begin{abstract}
The AlphaZero framework provides a standard way of combining Monte Carlo planning with prior knowledge provided by a previously trained policy-value neural network. AlphaZero usually assumes that the environment on which the neural network was trained will not change at test time, which constrains its applicability.
In this paper, we analyze the problem of deploying AlphaZero agents in potentially changed test environments and demonstrate how the combination of simple modifications to the standard framework can significantly boost performance, even in settings with a low planning budget available. The code is publicly available on GitHub\footnote{\url{https://github.com/TheEmotionalProgrammer/az-generalization}}.

\end{abstract}
\section{Introduction}
Imagine you are sitting on the passenger seat of your brand-new self-driving car. Your car is equipped with a navigator, used to dynamically plan the most convenient path to your destination; of course, exploring every possible path is unfeasible. To guide its planning, the car runs a neural network under the hood, capable of directing the path search with a standard AlphaZero (AZ) \cite{AlphaZero} procedure, using the navigator as the planning model. This neural network was accurately trained on high-fidelity traffic simulations a few years ago, but over time, the topology of your neighborhood changed and does not fully reflect the original city environment anymore, which the neural network expects. One day, you encounter an unexpected road closure on your most convenient path to the destination; the car starts computing an alternative best path, but the neural network is just too overfitted on the original topology. As a result, the car realizes that it has to steer when it is too late, leading you to crash into a wall. Fortunately, your car is equipped with a nice airbag, but repairing the damage will cost a good portion of your yearly salary.

The situation just described highlights the potentially problematic application of modern model-based reinforcement learning algorithms, such as AlphaZero, when the test environments the agent is deployed to can differ from training, which invalidates the neural network predictions. On one hand, completely disregarding the network in favor of an unbiased Monte Carlo Tree Search (MCTS) \cite{remi,kocsis} would discard a large amount of useful information that practically makes it feasible to plan on the go in complex environments. On the other hand, planning with imperfect estimators could lead to dangerous situations where the online search cannot account quickly enough for the wrong prior beliefs of the neural network. Therefore, finding a way to make these algorithms robust to deployment-time variations is fundamental for their application outside of controlled, simulated settings.

In this paper, we analyze the problem of planning with learned estimators in a partially modified test environment, with a particular focus on the AlphaZero framework. In doing so, we underline potential weaknesses and inefficiencies of the original algorithm, review existing MCTS techniques that allow for better use of the available planning budget, and propose to combine some of them in a novel algorithm called Extra-Deep Planning (EDP). 

The EDP algorithm allows the agent to build increasingly deep trees during planning, which manage to more quickly adjust the wrong predictions of the neural network and therefore switch the agent's focus to paths that are optimal at test time. Moreover, we develop a simple set of grid-world experiments that demonstrate EDP's superior performance compared to standard AZ and allow for a comprehensive analysis of the contribution of each algorithm's component to such improvements. We find that while the individual components yield only modest benefits on their own, their combination makes the agent act nearly optimally in navigation tasks even with a low planning budget.
\section{Background}
\subsection{Reinforcement Learning}

Reinforcement Learning (RL) is a branch of machine learning concerned with how agents take actions in an environment to maximize cumulative reward. It formalizes learning from interaction through feedback in the form of rewards or penalties. The standard framework for RL is the Markov Decision Process (MDP)~\cite{bellman1957markovian}, defined as a tuple:
\[
(\mathcal{S}, \rho, \mathcal{A}, \mathcal{P}, \mathcal{R}, \gamma),
\]
where \( \mathcal{S} \) is the set of states, \( \rho \) the initial state distribution, \( \mathcal{A} \) the set of actions, \( \mathcal{P}(s' | s, a) \) the transition probability, \( \mathcal{R}(s, a, s') \) the reward function, and \( \gamma \in (0,1] \) the discount factor.

The objective in RL is to find a policy \( \pi \) that maximizes the expected return, defined as the cumulative discounted reward:
\[
V_\pi(s) = \mathbb{E} \left[ \sum_{t=0}^{\infty} \gamma^t r_t \,\middle|\; \stack{s_0 = s,\; a_t \sim \pi(s_t) \brs s_{t+1} \sim \mathcal P(s_t,a_t) \brs r_t = \mathcal R(s_t, a_t, s_{t+1})}\, \right],
\]
with associated action-value function, also known as one of the Bellman equations \cite{bellman1957dynamic}:
\[
Q_\pi(s, a) = \mathbb{E} \left[ \mathcal R(s,a,s') + \gamma V_\pi(s') \,\middle|\, s' \sim \mathcal P(s,a) \right].
\]

In deterministic environments, both the transition and reward functions simplify to deterministic mappings \( s_{t+1} = f(s_t, a_t) \), \( r_t = \mathcal R(s_t, a_t, f(s_t, a_t)) \). These assumptions apply throughout this paper unless stated otherwise.

RL methods are typically grouped into two categories:
\begin{itemize}
    \item \textbf{Model-free methods:} These learn directly from experience without access to the transition model. They include Value-based methods, such as Q-learning~\cite{qlearning} and Deep Q-Networks~\cite{mnih}, and  Policy-based methods, such as REINFORCE~\cite{REINFORCE} and Actor-Critic algorithms~\cite{acmethods}.
    \vspace{2mm}
    \item \textbf{Model-based methods:} These incorporate or learn a model of the environment to simulate and evaluate future trajectories. Notable examples include MCTS~\cite{remi,kocsis}, AlphaZero~\cite{AlphaZero}, and MuZero~\cite{MuZero}.
\end{itemize}
Model-free methods are generally simpler and avoid the problems of model bias and error, but require direct interaction with the environment during learning.
Model-based methods, on the other hand, tend to be more sample-efficient at the cost of greater computational complexity during runtime.

Given the particular relevance of AlphaZero in the context of this paper, we devote the next \autoref{subsec-az-framework} to explaining its fundamentals.
\begin{figure}[b!]
    \centering
    \includegraphics[width=\linewidth]{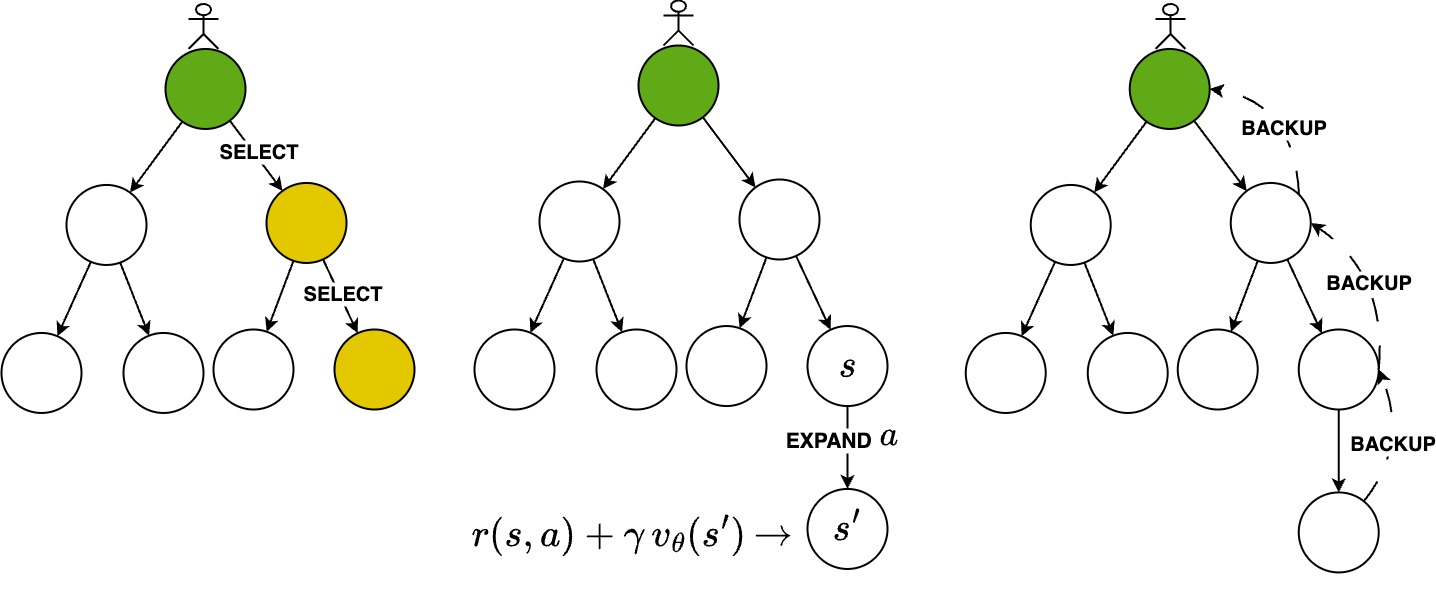}
    \caption{A single iteration of AZ planning, from left to right: the agent selects a path using a selection policy (e.g., PUCT \eqref{PUCT}) until reaching a not-fully-expanded node, expands it through a sampled action, and estimates the value of the newly created node using the value network. Its value is then backpropagated through the path until reaching the root, updating the mean value estimates along the path. This procedure is repeated for $B$ iterations, where $B$ is the planning budget.}
    \label{fig:az_planning_viz}
\end{figure}
\subsection{The AlphaZero Framework}
\label{subsec-az-framework}
AlphaZero (AZ) \cite{AlphaZero} extends standard Monte Carlo Tree Search planning (MCTS) \cite{remi,kocsis} by incorporating a policy-value neural network $f_\theta: \mathcal S \rightarrow {\Delta} \times \mathbb R $ parametrized by $\theta$. which takes a state $s$ as an input and outputs a policy distribution $\pi \in \Delta := [0,1]^{|\mathcal A|}, \|\pi \| = 1$ and a value $v \in \mathbb R$. For notation convenience, we will indicate the value-head output given input $s$ as $v_\theta(s)$ and the corresponding policy-head output as $\pi_\theta(s)$.

The AZ planning procedure is visualized in \autoref{fig:az_planning_viz}, where each node in the planning tree represents a unique sequence of actions starting from the root node\footnote{Directly defining nodes as states would be restrictive and could cause issues in stochastic or cyclic
environments.}, and the root node corresponds to the current state in the environment.

To grow a planning tree, the AZ agent iteratively traverses a path of the tree built so far according to a selection policy. Once it reaches a node that is not fully expanded, an action is sampled among the non-expanded ones to create a new node, and the value of the corresponding state is estimated using the value network. Finally, the value is backed up through the path to update the parents' value estimates. This procedure can be repeated for a fixed number of times $B$, usually called the planning budget.

The most common selection policy used to traverse the tree is based on the PUCT formula \cite{puct} that assigns a score to selecting action $a$ from node $x$ containing state $s$ as:\begin{equation}
\text{PUCT}(a|x) = \bar{Q}(x \uplus a) + C \, \pi_\theta(a|x) \, \frac{\sqrt{N(x)}}{1 + N(x \uplus a)}
\label{PUCT}
\end{equation}
where:
\begin{itemize}
    \item $x \uplus a$ is the node we enter by selecting action $a$ from node $x$.
    \item $\bar{Q}(x \uplus a)$ is the current mean value estimate of the node $x \uplus a$.
    \item $\pi_\theta(a|x) = \pi_\theta(a|s)$ and $\pi_\theta(x) = \pi_\theta(s)$ for state $s$ corresponding to node $x$.
    \item $N(x)$ is the total visitation count of node $x$.
    \item $C$ is an exploration hyperparameter.
\end{itemize}
The mean value estimate $\bar{Q}(x)$ is computed as the average value of each planning trajectory that has passed through $x$ so far, resulting in the following recursive formulation:
\begin{equation*}
    \bar{Q}(x) = r(x) + \gamma\,\frac{v_\theta(x)}{N(x)} +  \gamma \sum_{a \in \mathcal A}\frac{N(x \uplus a)} {N(x)}\bar{Q}(x \uplus a)
\end{equation*}
Intuitively, PUCT balances exploitation of actions with high estimated value and exploration of actions favored by the policy network but visited less often.

Note that the formula \eqref{PUCT} originates from the common UCT formula \cite{kocsis} used in standard MCTS, which is in turn based on the Upper Confidence Bound algorithm (UCB) for multi-armed bandits \cite{ucb}.
In UCT, the exploration term is not guided by any prior policy:
\begin{equation}
    \text{UCT}(a|x) = \bar{Q}(x \uplus a) + C \, \sqrt\frac{\log N(x)}{N(x \uplus a)}
    \label{UCT}
\end{equation}
Once the planning tree is constructed, the agent needs to choose an action to undertake in the real environment. This is done according to an evaluation policy which is usually based on the visitation counts at the root $x_0$ after planning:
\begin{equation}
    \pi_N(a | x_0) = \frac{N(x_0\uplus a)}{N(x_0)}
    \label{visitcounts_evaluator}
\end{equation}
This can be turned into a deterministic policy by simply taking its argmax.

\subsubsection{Training}
In principle, it would be possible to deploy an agent in an environment using AZ planning given any policy and value neural network(s) trained on the same environment, regardless of the way training was performed (e.g., by using a PPO \cite{schulman2017proximalpolicyoptimizationalgorithms} agent). However, AZ also provides a standard way of training the network by following a self-play loop alternating between data collection, network updates, and periodic evaluation. Since knowing the details of the AZ training process is not necessary for understanding this paper, we now briefly summarize the process and provide more details in the appendix (\autoref{app: training}).

In the data collection phase, the agent interacts with the environment using the evaluation policy $\pi_\text{eval}$ derived from planning (typically $\pi_\text{eval} = \pi_N$). At each state, the mentioned planning procedure provides both an improved policy distribution $\pi_t$ and a value estimate, and the resulting sequences of states, actions, rewards, and policies are stored in a replay buffer.

During the network update phase, minibatches from the replay buffer are used to optimize the policy-value network $f_\theta$. The loss combines a value loss, aligning predictions with $n$-step return targets \cite{vanSeijen2015}, and a policy loss encouraging the network to match the improved search policies obtained from planning \cite{AlphaGo}.

Finally, in the evaluation phase, the agent’s performance is periodically assessed using a deterministic version of the evaluation policy.

\section{Methodology}
The occurrence of local changes to the test environment implies that the available $\pi_\theta$ and $v_\theta$ will not be fully accurate. However, given an arbitrarily high planning budget, AZ planning would eventually account for these errors. The problem is, therefore, how to effectively exploit the available budget by leveraging the information provided by the neural network without overcommitting to it.

In the remainder of this section, we will assume a deterministic and fully observable environment. Moreover, we will assume that the environment can change at test time compared to the training setting, but does not change further during deployment, i.e., the test environment is stationary.

\subsection{The Exploration-Exploitation Trade-off}
\label{subsec: exploration}
As described in \autoref{subsec-az-framework}, the $C$ parameter regulates how greedy we want to be during planning. Intuitively, we might think that changes to the environment require more exploration (higher $C$). However, this exploration is also guided by the neural network in AZ, since we are using PUCT \eqref{PUCT}. We could therefore switch back to using standard UCT \eqref{UCT} with a uniform exploration to avoid being fooled by the policy network, but the resulting planning tree still depends on the learned value function and additionally exhibits a high branching factor that prevents the agent from looking far into the future. Moreover, checking a certain path a small number of times does not allow the agent to properly update the corresponding estimates. To better visualize this, one could imagine an agent bumping into a closed door that was open during training. Repeatedly bumping into the door would slowly decrease the value in front of it due to discounting, even if bumping does not yield a negative reward. But it would take a lot of tries to see a significant difference, as wrongly estimated values can only be reduced by discounting. If we can plan for enough time, it might therefore be better to let the agent be greedy w.r.t.\ its estimates ($C \approx 0$) so that the resulting deeper tree can identify the changes faster and plan around them. 

\subsection{Tree Recycling}
\label{subsec: tree-reuse}
\begin{figure}[t!]
    \centering
    \includegraphics[width=\linewidth]{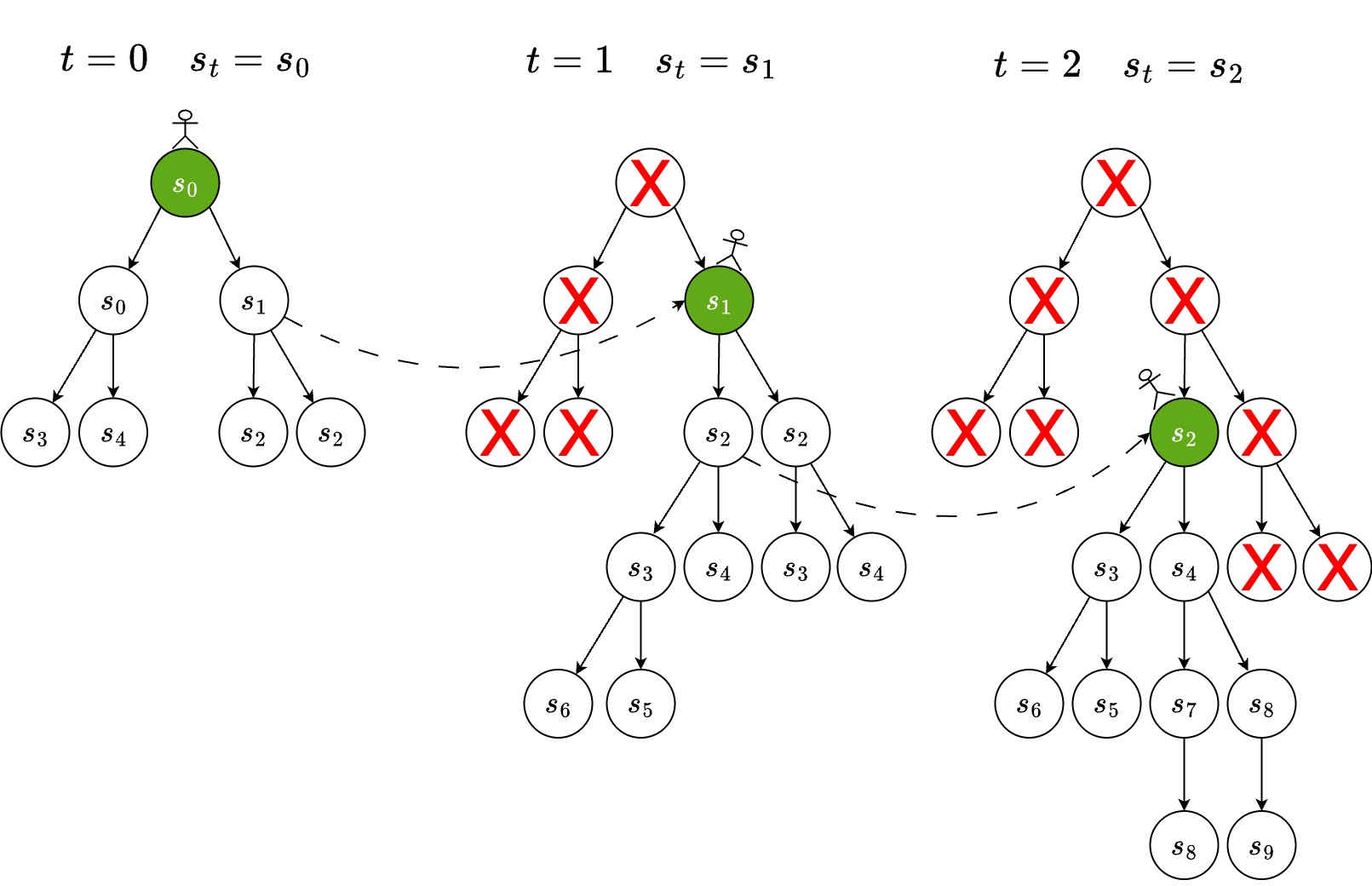}
    \caption{Example of the tree re-usage mechanism. The green node in each tree corresponds to the current state $s_t$ in the real environment at step $t$. After step $1$, we can reuse the right subtree of the previous root node as the corresponding child is the only one whose state is $s_1$ (the current state in the real environment). The left subtree is discarded (red-crossed nodes in the figure). After step $2$, we could reuse both children of the previous root, but we choose the left one as the corresponding subtree is deeper.}
    \label{fig:tree-reuse}
\end{figure}
Although being greedy during planning can help us better realize that something is off in the environment, we are still strongly constrained by the available planning budget. In standard AZ implementations, the planning tree is initialized from scratch after every step in the real environment. There are some key reasons why this is a common practice in the literature:\begin{itemize}
    \item \textbf{Model inaccuracy}: If our transition model is inaccurate, we might accumulate planning errors and end up modeling highly inaccurate transitions. 
    \item \textbf{Memory and time constraints}: Building an incrementally deeper tree requires recursively traversing and backing up from much longer trajectories, which might be undesirable if there are specific time constraints. It also requires more memory to maintain the increasing number of nodes\footnote{For sufficient tree exploration the average memory complexity of tree recycling is bounded. If the recycled subtree has complexity $\mathcal O(B/|\mathcal A|)$ for a planning budget $B$, then after $n$ steps in the environment the memory complexity of the recycled tree will be $\mathcal O(B^*_n) = \mathcal O(B + \frac{1}{|\mathcal A|} B^*_{n-1}) = \mathcal O(B \sum_{t=0}^{n-1} \frac{1}{|\mathcal A|^t})$, that is, $\lim\limits_{n \to \infty} \mathcal O(B^*_n) = \mathcal O(B \frac{|\mathcal A|}{|\mathcal A|-1}).$}.
\end{itemize}
While we acknowledge these limitations, we still believe that allowing our agent to reuse part of the previous planning tree can significantly reduce the amount of planning budget needed to properly plan in modified environments, particularly for the much deeper planning trees built by a greedy selection policy. Moreover, specific memory and/or recursion limits could simply be addressed by setting a maximum number of consecutive tree reuses, after which we can drop the previously constructed tree and start planning from scratch.

One way of reusing the previous tree is to check whose child of the previous root node holds the same state as in the real environment (if any) and resume planning from there. Note that there might be multiple children of the root that correspond to the current state, and in that case, we should then decide which subtree we want to retain. Since our main goal is building increasingly deep trees, we can choose the child whose corresponding subtree is the deepest, where the depth of a node's subtree is defined recursively as the "height" of the node:
\begin{equation*}
    \mathrm{height}(x) = 1 + \max_{x' \in \mathrm{children}(x)} \mathrm{height}(x')
\end{equation*}
\noindent
The recursion stops at leaf nodes whose height is set to 0 by definition. The tree re-usage mechanism is exemplified in \autoref{fig:tree-reuse}.

\subsection{Blocking Loops}
\label{subsec: block-loops}
\begin{figure}
    \centering
    \includegraphics[width=0.8\linewidth]{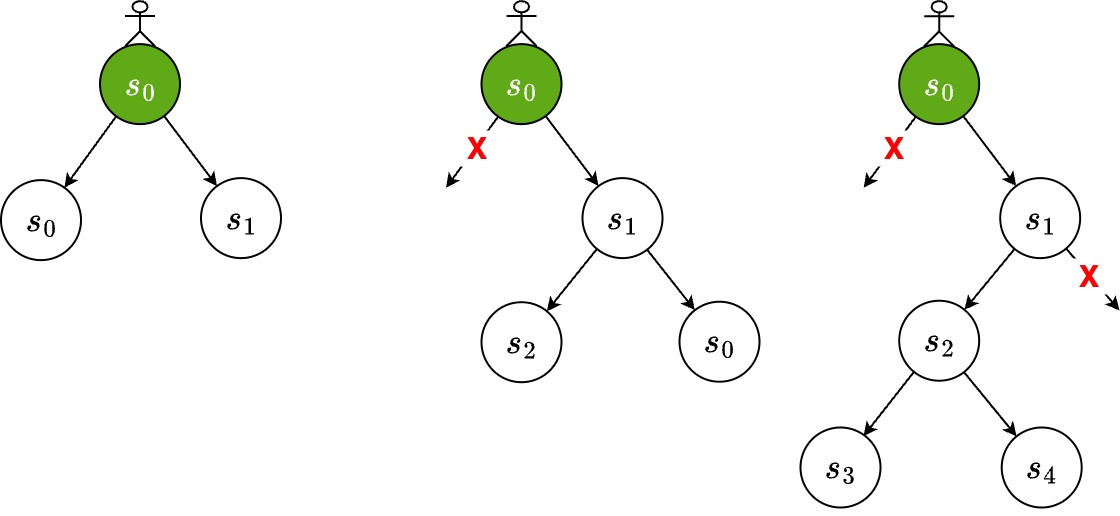}
    \caption{Example of loop blocking mechanism. During traversal, we prune the edges of the tree leading to states that had already been visited along the path.}
    \label{fig:block-loops}
\end{figure}

Another source of inefficiency of AZ planning that can be particularly detrimental when changes happen is the fact that we might explore the same path over and over because the (in the test environment) incorrectly estimated value for an action is comparatively high, entering a planning loop. Formally, we define a loop as the repetition of the same state along a single path from root to leaf.

Moerland et al. \cite{moerland} tested the possibility of blocking the loops in MCTS planning by setting the variance of a repeating state along a planning path to zero. In standard AZ planning, there is no such concept as the "variance" of a node\footnote{In theory, a variance of zero would correspond to an infinitely visited node.}. We instead propose to block loops by directly pruning the parent action from the search tree and making it not selectable from then on, as illustrated in \autoref{fig:block-loops}. To check whether a state is a loop, it is sufficient to compare it with every state above it, for example, during the back-up phase or by maintaining a set of the visited states during traversal\footnote{Checking whether a newly created (leaf) node already exists along the corresponding root-to-leaf path costs $\mathcal O(h)$ operations where $h$ is the height of the tree, i.e.\ $\mathcal O(\log_{|\mathcal A|} N)$ for roughly balanced trees with branching $|\mathcal A|>1$ and $N$ total nodes, and up to $\mathcal O(N)$ in the worst case when the tree is not balanced. However, by maintaining a path hash set, we get expected $\mathcal O(1)$ per node using $\mathcal O(h)$ memory.}.

In a deterministic environment, blocking the loops should be enough to avoid exploring over and over redundant paths that make us waste planning budget. For example, an agent in front of a newly introduced obstacle might spend a lot of planning time stuck before understanding that it should change direction.

While checking for exact state loops only applies to discrete state spaces, it is possible to extend this to continuous state spaces by checking that the norm between the states $s$ and $s'$ that we compare is below a threshold $\eta$, i.e. $\|s - s'\|_2 \leq  \eta, \,\,\, \eta \in \mathbb R$ for any pair of nodes $s,s'$ along the planning path. Setting $\eta=0$ corresponds to only blocking exact loops as we do in the discrete case.

To summarize, our modifications to standard AZ are the following: \begin{itemize}
    \item Greedy planning by setting $C = 0$.
    \item Reusing the previous planning tree at each step as described in \autoref{subsec: tree-reuse}.
    \item Blocking planning loops as described in \autoref{subsec: block-loops}.
\end{itemize}
Note that the action executed in the environment after planning is still chosen according to the standard AZ tree evaluation policy \eqref{visitcounts_evaluator}.

The combination of AZ and the described additional features constitutes the Extra-Deep Planning algorithm (EDP), and ablations are shown in \autoref{fig:ablation}. 

\begin{figure}[t!]
    \centering
    \includegraphics[width=\linewidth]{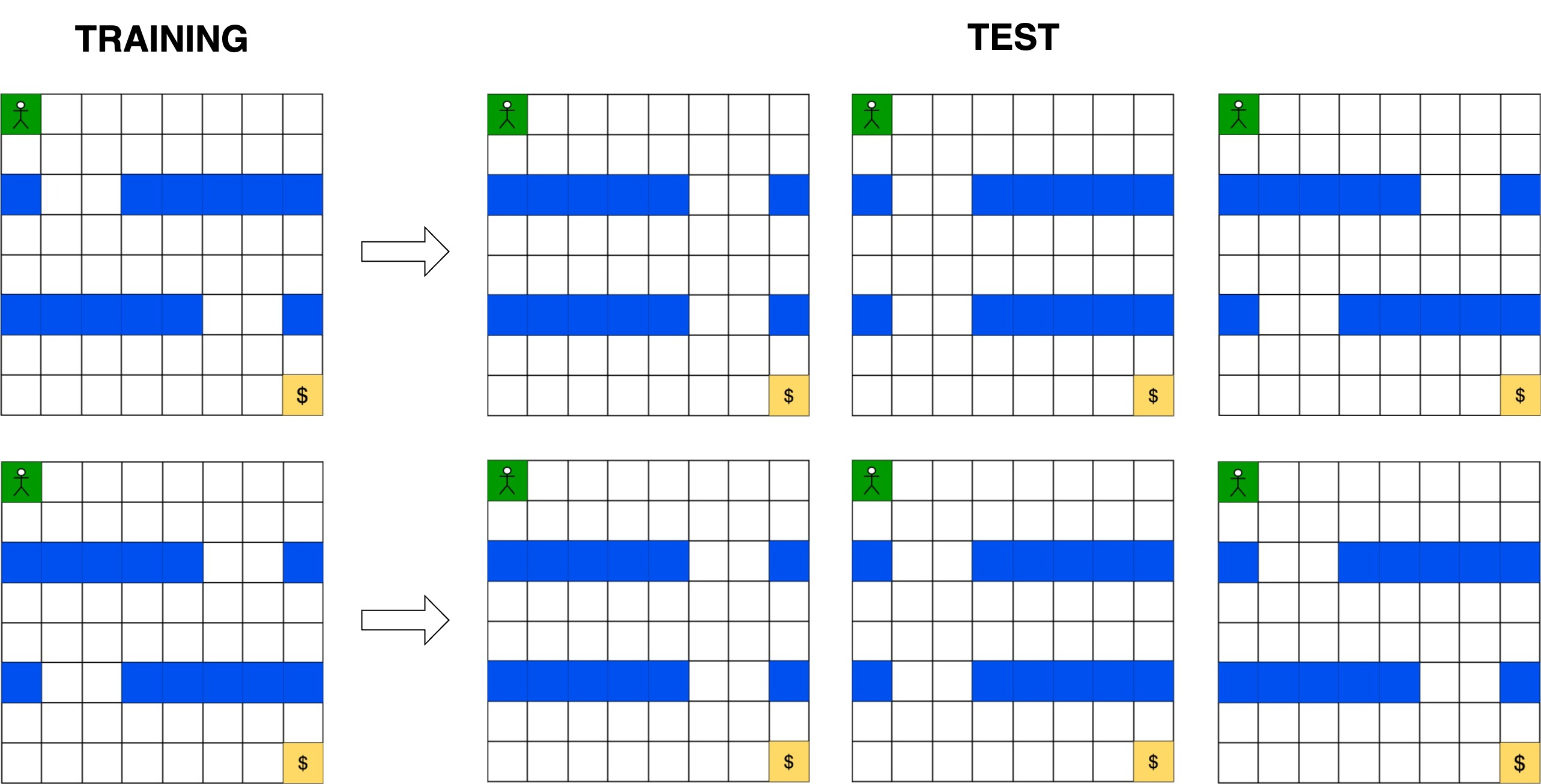}
    \caption{Overview of the MAZE training and test configurations. The green square represents the agent's starting position, the blue squares represent obstacles, and the yellow square represents the goal position. The agent is trained on the MAZE\_LR and MAZE\_RL configurations and can be tested by moving the holes in the walls into different positions, as shown in the figure.}
    \label{fig:challenges-mazes-viz}
\end{figure}

\subsection{Experimental Setup}
\label{subsec: setup}
We can now detail the employed environment configurations and agents for our evaluation.

\subsubsection{Environment}
To validate the effect of environment changes on AZ's performance, we propose a simple set of MAZE grid-world environments where the test configuration differs from the training configuration. These challenges are represented in \autoref{fig:challenges-mazes-viz}. As shown in the figure, all the configurations are $8\times8$ grids where the agent always starts from the upper-left corner $(0,0)$ (green location) and needs to reach the bottom-right corner $(7,7)$ (yellow location) by finding its way beyond the obstacles (blue locations). From now on, we will use the writing convention MAZE\_$X$ $\rightarrow$ MAZE\_$Y$ to indicate training on the MAZE\_$X$ configuration and testing on the MAZE\_$Y$ configuration, where $X$ and $Y$ can be LR (Left-Right), RL, LL, or RR depending on the position of the holes in the two horizontal walls. The agent can move left, right, up, or down (no diagonal move), and blue obstacles block movement so that the agent remains in the same state without incurring any punishment.

\subsubsection{Agents}
We evaluate the performance of our novel EDP algorithm against the following baselines:
\begin{itemize}
    \item Standard AlphaZero (AZ+PUCT): this is the baseline agent also used during training, which relies on the PUCT formula for node selection. 
    \item AlphaZero without prior policy (AZ+UCT): we consider this an equally relevant baseline because, depending on how much the test configuration diverges from the training environment, avoiding reliance on the prior policy might already lead to improved results. 
\end{itemize}
Details on the performed tuning of the PUCT/UCT $C$ parameter for the AZ baselines and on how the neural networks were trained can be found in the appendix (\autoref{app: tuning} and \autoref{app: training}, respectively).

\section{Related Work}
As previously discussed, AlphaZero has traditionally been applied to environments that do not change at test time, and the specific challenges analyzed in this paper have received, to the best of our knowledge, limited attention in the literature.

Min and Motani \cite{bricktictactoe} introduced Brick Tic-Tac-Toe, a new benchmark extending the classic Tic-Tac-Toe game, to evaluate AlphaZero in novel test configurations. Their findings show that training on diverse configurations improves test-time adaptation, but their setup differs from ours, as it involves a two-player game, while we focus on single-agent environments. Moreover, they do not propose novel planning methods addressing the challenge. 

\begin{figure}[t!]
    \centering
    \includegraphics[width=\linewidth]{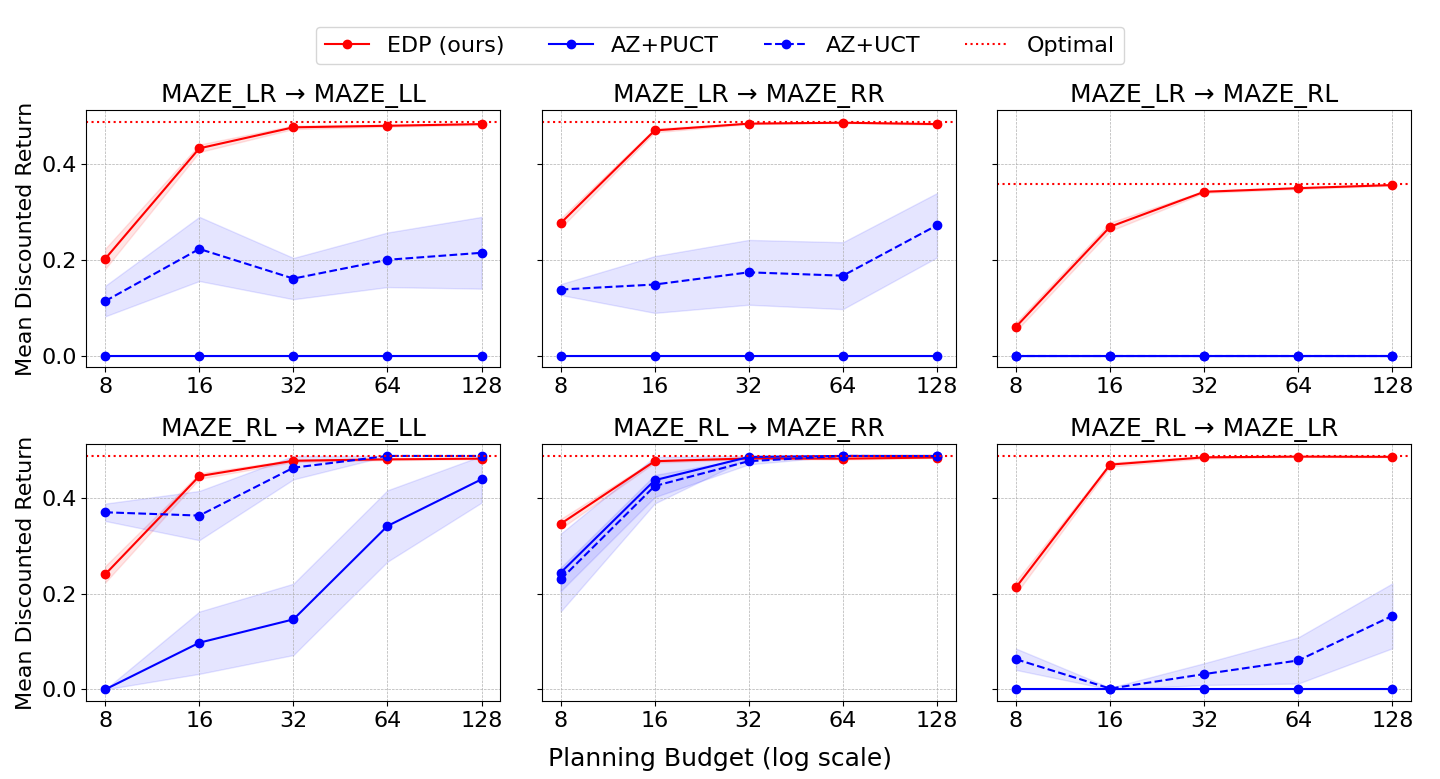}
    \caption{Results of the EDP planning algorithm (red line) on the MAZE grid-world challenges, compared with standard AZ (blue lines). Straight and dashed lines are reported for AZ+PUCT and AZ+UCT versions, respectively. The displayed metric is the mean discounted return averaged across $10 \times10$ training/evaluation seeds, and the shaded area represents the standard error across training seeds. The label MAZE\_$X$ $\rightarrow$ MAZE\_$Y$ on top of each plot indicates training on the MAZE\_$X$ configuration and testing on MAZE\_$Y$.}
    \label{fig:main-results}
\end{figure}

Lan et al. \cite{alphazerolikeagentsrobustadversarial} assess AlphaZero’s robustness to adversarial state perturbations in Go, though their perturbation model is domain-specific and does not involve changes to the algorithm itself. 

Pettet et al. \cite{tuedm} propose a novel MCTS-based algorithm addressing non-stationary environments called Policy-Augmented MCTS (PA-MCTS), combining a previously learned Q-function with online MCTS estimates. Unlike our approach and AlphaZero, which integrate prior knowledge within the planning tree, PA-MCTS applies it externally and weights the online and prior knowledge using a tunable hyperparameter.

\section{Results}
\label{sec:results}

We employ the discounted return $G_t$ as the primary metric. In our case, the reward function only returns a positive reward of 1 when reaching the goal; thus, this can simply be defined as:
\[
    G_t =
    \begin{cases}
        \gamma^{t} & \text{if goal is reached}  \\
        0 & \text{otherwise}
    \end{cases}
    \]
where $t$ is the number of steps it took the agent to reach the goal. The discount factor $\gamma=0.95$ is kept identical to the corresponding value during training (see \autoref{app: training}). If the agent has not reached the goal after 100 steps, then the episode is terminated with a reward of zero.

\begin{figure}[t!]
    \centering
    \includegraphics[width=\linewidth]{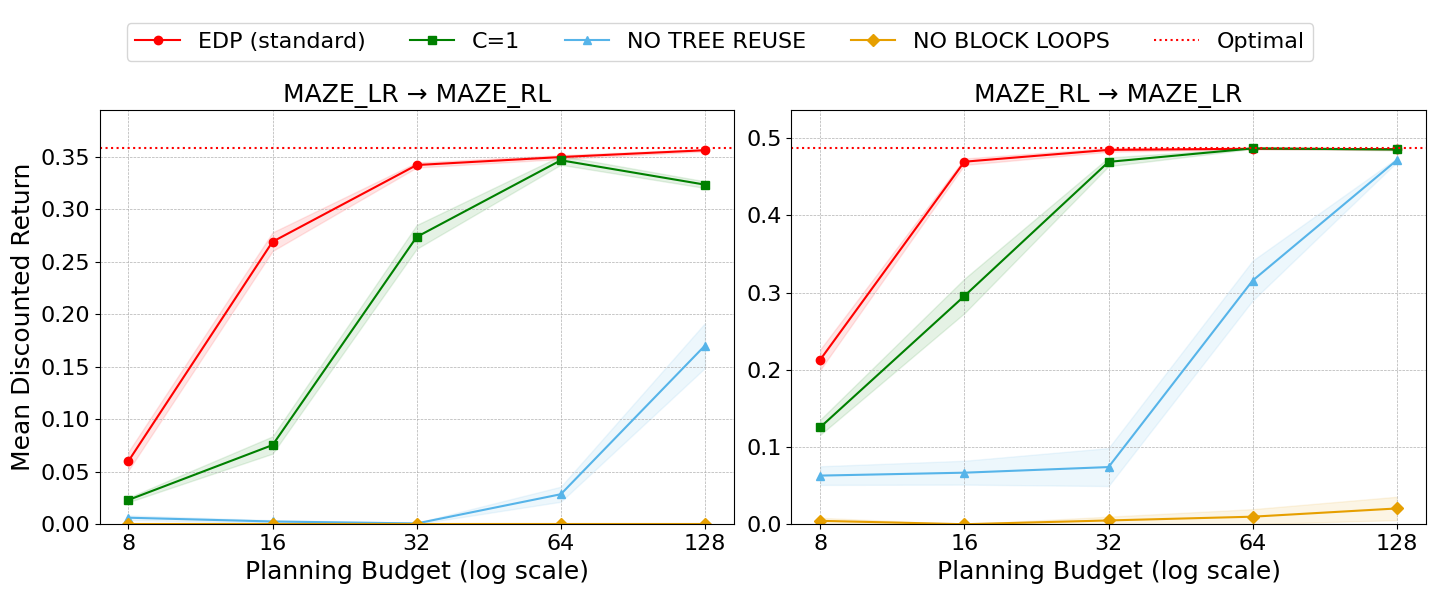}
    \caption{Ablation study of main EDP features of MAZE\_LR $\rightarrow$ MAZE\_RL and MAZE\_RL $\rightarrow$ MAZE\_LR challenges. The displayed metric is the mean discounted return averaged across $10 \times10$ training/evaluation seeds, and the shaded area represents the standard error across training seeds. The red line represents the standard EDP algorithm. The green line represents EDP with added UCT planning exploration, i.e., $C=1$ instead of $C=0$. The cyan line represents EDP without reusing the previous planning tree at each step. The dark yellow line represents EDP without blocking loops.}
    \label{fig:ablation}
\end{figure}

 \autoref{fig:main-results} shows the mean discounted return averaged across $10\times10$ training/evaluation seeds (10 evaluation seeds for each of 10 training seeds) for all the test configurations previously described and shown in \autoref{fig:challenges-mazes-viz}. In general, it is easy to see how EDP consistently outperforms the baselines on all the configurations.
Although removing the prior policy network improves the performance of the baseline AZ in some of the experiments (e.g., MAZE\_LR $\rightarrow$ MAZE\_LL and MAZE\_LR $\rightarrow$ MAZE\_RR), that remains vastly lower than EDP's.

\begin{figure}[t!]
    \centering
    \subfloat[Planning without blocking loops.\label{fig-a-loop}]{
        \includegraphics[width=0.45\linewidth]{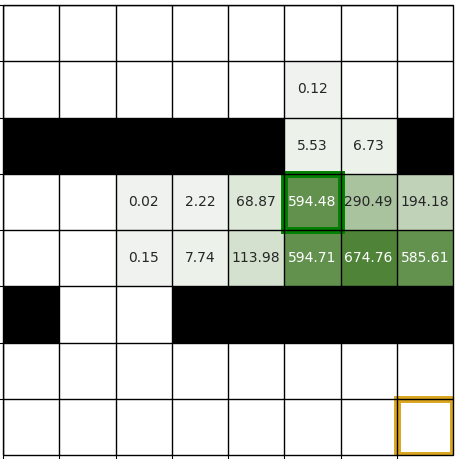}
    }
    \hfill
    \subfloat[Planning by blocking loops.\label{fig-b-loop}]{
        \includegraphics[width=0.45\linewidth]{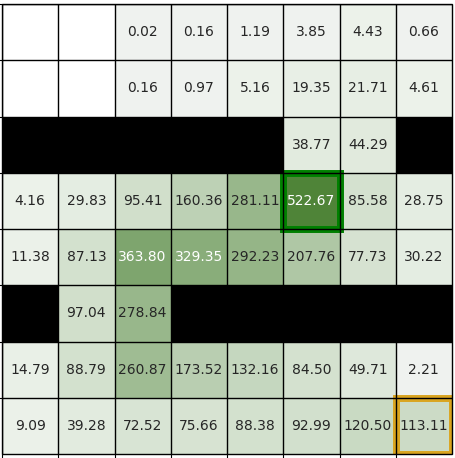}
    }
    \caption{Comparison of the state visitation counts of EDP planning with and without blocking loops on the MAZE\_LR $\rightarrow$ MAZE\_RL test. The agent is positioned in (3,5) (green border state) and builds a large planning tree of 512 nodes. The goal state (7,7) is circled in gold. The displayed values are the average visit counts of each state across $10\times10$ training/evaluation seeds. Note that the displayed numbers are not equivalent to the visitation counts for each node, since the tree can have several nodes containing the same state along a single path.}
    \label{fig:exp-loop}
\end{figure}
What makes these results particularly impressive is the performance difference between standard AZ and EDP on the "inverted" test configurations, i.e., MAZE\_LR $\rightarrow$ MAZE\_RL and MAZE\_RL $\rightarrow$ MAZE\_LR, where the original optimal policy at training time is completely compromised. In both cases, AZ performance does not seem to pick up even with a relatively large planning budget, while EDP solves both configurations even with a small budget. 

To better investigate why this happens and which of the components of EDP make it that much more effective, we conduct an ablation study on these two test configurations; specifically, we test the standard EDP against three modifications:
\begin{itemize}
    \item C=1: we use a large UCT exploration bonus instead of zero.
    \item NO TREE REUSE: we remove the tree-reuse feature detailed in \autoref{subsec: tree-reuse}.
    \item NO BLOCK LOOPS: we remove the loop blocking feature detailed in \autoref{subsec: block-loops}.
\end{itemize}
The results of this ablation are reported in \autoref{fig:ablation}. As one can see, all the main features of EDP are important for its performance, as demonstrated by the standard algorithm achieving the best results in both cases. However, the performance drop given by removing them differs significantly among the ablated features. Increasing the planning exploration (C=1) significantly affects performance for low planning budgets but is less detrimental for high budgets. Removing tree reuse damages performance even more, although its effect also reduces when the budget increases. Most importantly, not blocking the loops drops the algorithm's performance (close) to zero regardless of the planning budget for both test configurations.

We further inspect the importance of blocking loops during planning by visualizing how the visitation of the state space changes when including or removing such a feature. \autoref{fig:exp-loop} shows the state visitation count distribution of an EDP agent building an extremely large planning tree of 512 nodes starting from state (3,5) in the MAZE\_RL configuration. We chose such a high planning budget because we want to see whether the agent not blocking loops will eventually be able to get around the obstacle and solve the challenge, something that seemed almost impossible with the budgets used in the ablation. This insight is confirmed in \hyperref[fig-a-loop]{Figure 7a}, where we can see the agent not blocking the loops getting stuck into planning towards the right corner, never managing to look beyond the obstacle. Conversely, the agent blocking loops in \hyperref[fig-b-loop]{Figure 7b} manages to spread its planning budget across the environment, see the goal state multiple times, and mostly focus its budget towards it.

\section{Conclusions}
In this paper, we analyzed the performance of AlphaZero algorithms when deployed in partially changed test environments and compared their performance with our novel EDP algorithm, which combines different modifications to the framework that improve the way we exploit the available planning budget. Our experiments demonstrate how the standard framework particularly struggles with the test configurations that significantly differ from training, i.e., the MAZE\_LR $\rightarrow$ MAZE\_RL and MAZE\_RL $\rightarrow$ MAZE\_LR challenges, where the optimal path that the agent should follow is inverted from training to test. In contrast, EDP reaches optimal performance in all the tested configurations. Our ablations confirm how all the novel components of the algorithm contribute to the performance gain, with blocking the loops being particularly crucial. 

An important extension to our work would be validating our approach on larger and more complex environments. It would be particularly important to do this in environments with continuous states, where the definition of a loop is less obvious and we would need to properly tune the threshold $\eta$. It would also be interesting to experiment on non-stationary environments that keep changing during test, where mechanisms like reusing the previous planning tree could be less effective or even detrimental if we do not properly incorporate information about the novel changes. This would address more realistic situations like traffic jams, where the agent needs to quickly adapt to the behavior of the other cars. 

Finally, our approach assumes a deterministic and fully-observable environment. In order to make the algorithm applicable to more complex real-world challenges, it would be important to extend the framework to deal with stochastic and partially observable environments.

\begin{credits}
\subsubsection{\ackname} We acknowledge the use of computational resources of the DelftBlue supercomputer, provided by Delft High Performance Computing Centre\\ (\url{https://www.tudelft.nl/dhpc}). 
This work was partially funded by the Dutch Research Council (NWO) project {\em Reliable Out-of-Distribution Generalization in Deep Reinforcement Learning} with project number OCENW.M.21.234. This research has also received funding from the KU Leuven Research Funds (C14/24/092).
\end{credits}

\subsubsection{\discintname}
The authors have no competing interests to declare that are relevant to the content of this article.

\bibliographystyle{splncs04}
\bibliography{main}

\begin{thebibliography}{10}
\providecommand{\url}[1]{\texttt{#1}}
\providecommand{\urlprefix}{URL }
\providecommand{\doi}[1]{https://doi.org/#1}

\bibitem{ucb}
Auer, P., Cesa-Bianchi, N., Fischer, P.: Finite-time analysis of the multiarmed bandit problem. Machine Learning  \textbf{47},  235--256 (05 2002). \doi{10.1023/A:1013689704352}

\bibitem{bellman1957dynamic}
Bellman, R.: Dynamic Programming. Princeton University Press, Princeton, NJ (1957)

\bibitem{bellman1957markovian}
Bellman, R.: A markovian decision process. Journal of Mathematics and Mechanics  \textbf{6}(5),  679--684 (1957)

\bibitem{gymopenai}
Brockman, G., Cheung, V., Pettersson, L., Schneider, J., Schulman, J., Tang, J., Zaremba, W.: Openai gym. \url{https://arxiv.org/abs/1606.01540} (2016), arXiv preprint arXiv:1606.01540

\bibitem{remi}
Coulom, R.: Efficient selectivity and backup operators in monte-carlo tree search. vol.~4630 (05 2006). \doi{10.1007/978-3-540-75538-8_7}

\bibitem{relu}
Glorot, X., Bordes, A., Bengio, Y.: Deep sparse rectifier neural networks. In: Proceedings of the Fourteenth International Conference on Artificial Intelligence and Statistics (AISTATS). pp. 315--323 (2011), \url{http://proceedings.mlr.press/v15/glorot11a/glorot11a.pdf}

\bibitem{adam}
Kingma, D.P., Ba, J.: Adam: A method for stochastic optimization (2017), \url{https://arxiv.org/abs/1412.6980}

\bibitem{kocsis}
Kocsis, L., Szepesv{\'a}ri, C.: Bandit based monte-carlo planning. In: F{\"u}rnkranz, J., Scheffer, T., Spiliopoulou, M. (eds.) Machine Learning: ECML 2006. pp. 282--293. Springer Berlin Heidelberg, Berlin, Heidelberg (2006)

\bibitem{acmethods}
Konda, V., Tsitsiklis, J.: Actor-critic algorithms. In: Solla, S., Leen, T., M\"{u}ller, K. (eds.) Advances in Neural Information Processing Systems. vol.~12. MIT Press (1999), \url{https://proceedings.neurips.cc/paper_files/paper/1999/file/6449f44a102fde848669bdd9eb6b76fa-Paper.pdf}

\bibitem{alphazerolikeagentsrobustadversarial}
Lan, L.C., Zhang, H., Wu, T.R., Tsai, M.Y., Wu, I.C., Hsieh, C.J.: Are alphazero-like agents robust to adversarial perturbations? (2022), \url{https://arxiv.org/abs/2211.03769}

\bibitem{bricktictactoe}
Min, J.T.C., Motani, M.: Brick tic-tac-toe: Exploring the generalizability of alphazero to novel test environments (2022), \url{https://arxiv.org/abs/2207.05991}

\bibitem{mnih}
Mnih, V., Kavukcuoglu, K., Silver, D., Graves, A., Antonoglou, I., Wierstra, D., Riedmiller, M.: Playing atari with deep reinforcement learning (2013), \url{https://arxiv.org/abs/1312.5602}

\bibitem{moerland}
Moerland, T.M., Broekens, J., Plaat, A., Jonker, C.M.: Monte carlo tree search for asymmetric trees (2018), \url{https://arxiv.org/abs/1805.09218}

\bibitem{tuedm}
Pettet, A., Zhang, Y., Luo, B., Wray, K., Baier, H., Laszka, A., Dubey, A., Mukhopadhyay, A.: Decision making in non-stationary environments with policy-augmented search (2024), \url{https://arxiv.org/abs/2401.03197}

\bibitem{puct}
Rosin, C.: Multi-armed bandits with episode context. Annals of Mathematics and Artificial Intelligence  \textbf{61},  203--230 (09 2010). \doi{10.1007/s10472-011-9258-6}

\bibitem{mlpandbackprop}
Rumelhart, D.E., Hinton, G.E., Williams, R.J.: Learning representations by back-propagating errors. Nature  \textbf{323}(6088),  533--536 (1986). \doi{10.1038/323533a0}

\bibitem{MuZero}
Schrittwieser, J., Antonoglou, I., Hubert, T., Simonyan, K., Sifre, L., Schmitt, S., Guez, A., Lockhart, E., Hassabis, D., Graepel, T., Lillicrap, T., Silver, D.: Mastering atari, go, chess and shogi by planning with a learned model. Nature  \textbf{588}(7839),  604–609 (Dec 2020). \doi{10.1038/s41586-020-03051-4}, \url{http://dx.doi.org/10.1038/s41586-020-03051-4}

\bibitem{schulman2017proximalpolicyoptimizationalgorithms}
Schulman, J., Wolski, F., Dhariwal, P., Radford, A., Klimov, O.: Proximal policy optimization algorithms (2017), \url{https://arxiv.org/abs/1707.06347}

\bibitem{AlphaGo}
Silver, D., Huang, A., Maddison, C.J., Guez, A., Sifre, L., van~den Driessche, G., Schrittwieser, J., Antonoglou, I., Panneershelvam, V., Lanctot, M., Dieleman, S., Grewe, D., Nham, J., Kalchbrenner, N., Sutskever, I., Lillicrap, T.P., Leach, M., Kavukcuoglu, K., Graepel, T., Hassabis, D.: Mastering the game of go with deep neural networks and tree search. Nature  \textbf{529},  484--489 (2016), \url{https://api.semanticscholar.org/CorpusID:515925}

\bibitem{AlphaZero}
Silver, D., Hubert, T., Schrittwieser, J., Antonoglou, I., Lai, M., Guez, A., Lanctot, M., Sifre, L., Kumaran, D., Graepel, T., Lillicrap, T., Simonyan, K., Hassabis, D.: Mastering chess and shogi by self-play with a general reinforcement learning algorithm (2017), \url{https://arxiv.org/abs/1712.01815}

\bibitem{gymnasium2023}
Towers, M., Terry, J.K., Kwiatkowski, A., Balis, J.U., de~Cola, G., Deleu, T., ao, M.G., Kallinteris, A., KG, A., Krimmel, M., Perez-Vicente, R., Pierr\'e, A., Schulhoff, S., Tai, J.J., Shen, A.T.J., Younis, O.G.: Gymnasium. \url{https://zenodo.org/record/8127025} (March 2023), zenodo

\bibitem{vanSeijen2015}
Vanseijen, H., Sutton, R.: A deeper look at planning as learning from replay. In: Proceedings of the 32nd International Conference on Machine Learning. Proceedings of Machine Learning Research, vol.~37, pp. 2314--2322. PMLR (2015)

\bibitem{qlearning}
Watkins, C.J.C.H., Dayan, P.: Q-learning. Machine Learning  \textbf{8}(3),  279--292 (1992). \doi{10.1007/BF00992698}, \url{https://doi.org/10.1007/BF00992698}

\bibitem{REINFORCE}
Williams, R.J.: Simple statistical gradient-following algorithms for connectionist reinforcement learning. Machine Learning  \textbf{8}(3),  229--256 (1992). \doi{10.1007/BF00992696}, \url{https://doi.org/10.1007/BF00992696}

\end{thebibliography}

\appendix
\section{Influence of the $C$ Exploration Parameter}
\label{app: tuning}
\begin{figure}[b!]
    \centering
    \includegraphics[width=\linewidth]{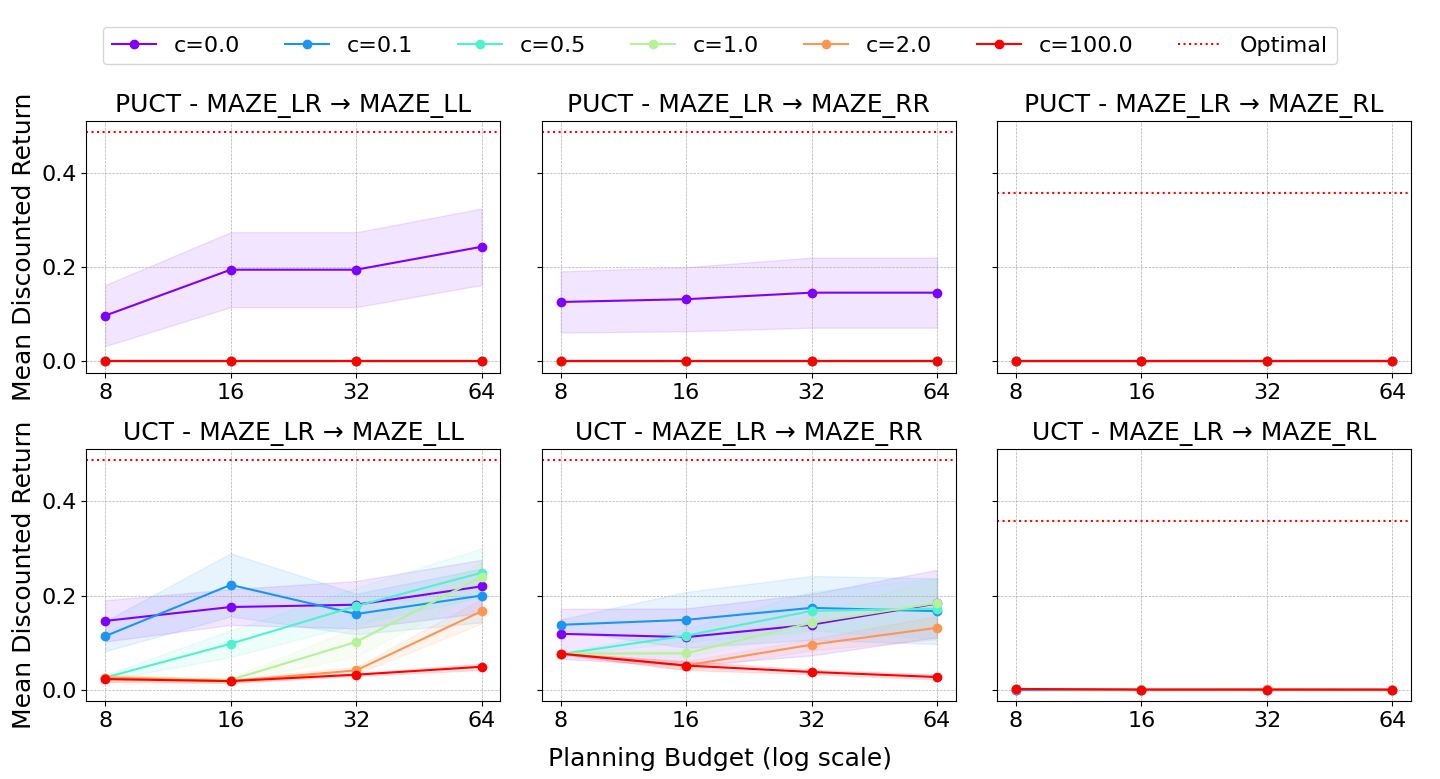}
    \caption{Influence of the $C$ parameter on AZ baselines in MAZE test configurations with MAZE\_LR training. A darker line color indicates a lower value of $C$. The label $POL$ - MAZE\_$X$ $\rightarrow$ MAZE\_$Y$ on top of each plot indicates training on MAZE\_$X$ and testing on the MAZE\_$Y$ with selection policy $POL$.}
    \label{fig:tuning_maze_lr}
\end{figure}
\begin{figure}[t!]
    \centering
    \includegraphics[width=\linewidth]{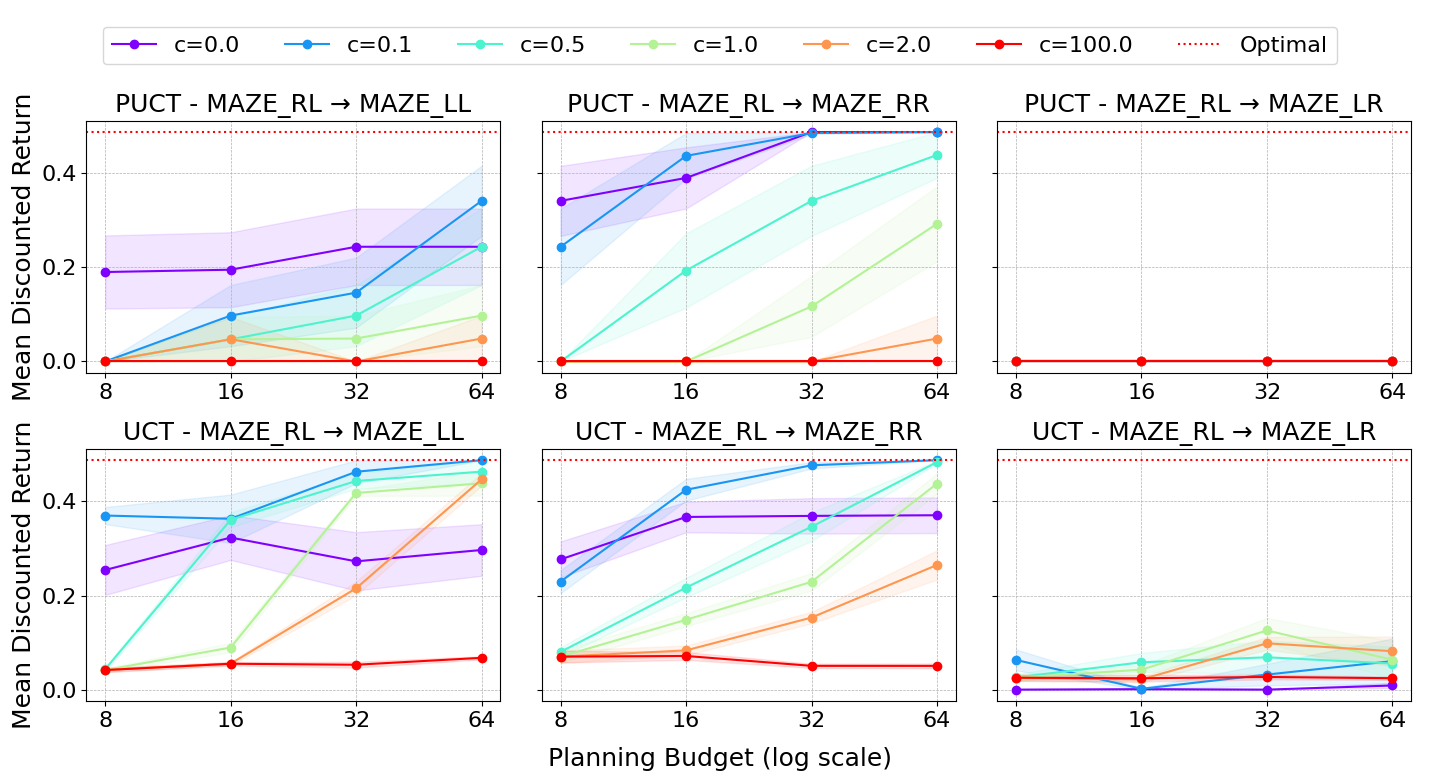}
    \caption{Influence of the $C$ parameter on AZ baselines in MAZE test configurations with MAZE\_RL training. A darker line color indicates a lower value of $C$. The label $POL$ - MAZE\_$X$ $\rightarrow$ MAZE\_$Y$ on top of each plot indicates training on MAZE\_$X$ and testing on MAZE\_$Y$ with selection policy $POL$.}
    \label{fig:tuning_maze_rl}
\end{figure}
The standard AZ planning only requires choosing the $C$ hyperparameter of PUCT/UCT at deployment, since the neural network is no longer updated. Note that the value for $C$ chosen during training does not necessarily represent the optimal choice at deployment. A higher value might be more useful during training, where exploration is strictly necessary, while a lower value at deployment can be a better choice if the prior estimates are reliable. As this is not necessarily true in our case due to the changed test configurations, we want to tune the parameter to compare our EDP algorithm to the best performance of the AZ baselines that we can achieve. We do this for both AZ+PUCT and AZ+UCT baselines:\begin{itemize}
    \item The tuning of $C$ for AZ+PUCT and AZ+UCT on MAZE test configurations with MAZE\_LR training is shown in \autoref{fig:tuning_maze_lr}.
    \item The tuning of $C$ for AZ+PUCT and AZ+UCT on MAZE test configurations with MAZE\_RL training is shown in \autoref{fig:tuning_maze_rl}.
\end{itemize}
We can see how varying $C$ influences the performance by a considerable margin, with the best results achieved with low values of the parameter (0, 0.1). This confirms our belief that being greedy can be surprisingly convenient when the test environment changes, as conjectured in \autoref{subsec: exploration}.

\section{Training Details and Hyperparameters}
\label{app: training}

\begin{table}[t!]
\centering
\begin{tabular}{|l|l|c|c|}
\hline
\textbf{Category} & \textbf{Parameter} & \textbf{MAZE\_LR} & \textbf{MAZE\_RL} \\
\hline

\multirow{2}{*}{Environment} 
    & \texttt{max\_ep\_len}      & 200 & 200 \\
    \cline{2-4}
  & \texttt{disc\_factor}       & 0.95 & 0.95 \\
\hline

\multirow{3}{*}{Training} 
  & \texttt{iterations}      & 100 & 150 \\
  \cline{2-4}
  & \texttt{learning\_epochs}      & 4 & 4 \\
  \cline{2-4}
  & \texttt{sample\_size}      & 6 & 6 \\
   \cline{2-4}
  & \texttt{buffer\_size}      & 90 & 90 \\
  \cline{2-4}
  & \texttt{batch\_size}      & 22 & 22 \\
  \cline{2-4}
  & \texttt{learning\_rate}      & 0.001 & 0.001 \\
  \cline{2-4}
  & \texttt{optimizer}      & Adam & Adam \\
   \cline{2-4}
  & \texttt{eval\_period}      & 10 & 10 \\

\hline

\multirow{2}{*}{Loss} 
  & \texttt{value\_weight}         & 0.7 & 0.7 \\
  \cline{2-4}
  & \texttt{policy\_weight}         & 0.3 & 0.3 \\
  \cline{2-4}
  & \texttt{n\_steps}      & 2 & 2 \\
\hline

\multirow{3}{*}{Neural Network} 
  & \texttt{hidden\_size}       & 64 & 64 \\
  \cline{2-4}
  & \texttt{hidden\_num}   & 2 & 2 \\
  \cline{2-4}
  & \texttt{activation} & ReLU & ReLU \\
\hline

\multirow{2}{*}{Planning} 
  & \texttt{planning\_budget}   & 64 & 64 \\
  \cline{2-4}
  & \texttt{c}       & 0.5 & 0.5 \\
  \cline{2-4}
   & \texttt{dir\_eps}      & 0.4 & 0.4 \\
   \cline{2-4}
  & \texttt{dir\_alpha}      & 2.5 & 2.5 \\
\hline

\end{tabular}
\vspace{1.0em}
\caption{Training parameters employed for the MAZE\_LR and MAZE\_RL grid-world environments.}

\label{tab:train-hyp-tab}
\end{table}

In this section, we detail our AZ training setting, which we used to obtain the trained policy-value neural network necessary for deploying both the baseline AZ agent and the novel EDP agent. The utilized hyperparameters are reported in \autoref{tab:train-hyp-tab}. Moreover, all the necessary source code for performing training, as well as our pre-trained weights, are publicly available at the associated \href{https://github.com/TheEmotionalProgrammer/az-generalization}{repository}.

The NN training is carried out by alternating three phases, with a full cycle constituting one of multiple \texttt{iterations}.

In the \textbf{sampling} phase, the agent collects \texttt{sample\_size} episodes by planning and acting in the environment using its current estimators $v_\theta$ and $\pi_\theta$. Actions are drawn from the (stochastic) evaluation policy $\pi_\text{eval}$, and Dirichlet noise is added to the root's prior policy logits to increase exploration. Episodes end at terminal states or after \texttt{max\_ep\_len} steps. No parameter updates occur; episodes are stored in a replay buffer of size \texttt{buffer\_size}. Each trajectory $T$ is stored as a tuple of sequences:
    \begin{equation}
            T = \left( \{s_t\}_{t=0}^{l}, \{a_t\}_{t=0}^{l-1}, \{r_t\}_{t=0}^{l-1}, \{\pi_t\}_{t=0}^{l-1} \right)
    \end{equation}
    where $l$ is the final timestep of the trajectory and $\pi_t = \pi_\text{eval}(s_t)$.

During \textbf{learning}, we run \texttt{learning\_epochs} epochs. At each epoch, we sample \texttt{batch\_size} episodes uniformly from the buffer to compute value and policy losses (details in \autoref{sec:losses_comp}). The \texttt{optimizer} (Adam \cite{adam}) updates the network parameters according to the chosen \texttt{learning\_rate}.

In the \textbf{evaluation} phase (run periodically, every \texttt{eval\_period} iterations), the policy is tested deterministically (argmax actions) without Dirichlet noise or parameter updates. We run evaluation episodes up to length \texttt{max\_ep\_len} and report average discounted and undiscounted returns. Training may be stopped early if performance is already near-optimal.
\subsection{Loss Computation}
\label{sec:losses_comp}
The AlphaZero loss is computed over a minibatch of $m$ trajectories $B = ( T_0, ..., T_m)$ sampled from the replay buffer as:
    \[
    L_{\text{\tiny{AZ}}}(B) = \alpha\, L_V(B) + \beta\, L_P(B)
    \]
     where $\alpha, \beta$ are weighting hyperparameters (\texttt{value\_weight} and \texttt{policy\_weight} in \autoref{tab:train-hyp-tab}), $L_V$ is the value loss and $L_P$ is the policy loss.
     
The value loss $L_V$ is defined as the mean squared error between the predicted state values and the $n$-step targets, averaged over all the $m$ trajectories in the minibatch:
\[
L_V(B) = \frac{1}{m} \sum_{j=1}^{m} \left\| \mathbf{v}_\theta(T_j[s]) - \mathbf{y}_n(T_j) \right\|^2
\]
where:
\begin{itemize}
    \item $\mathbf{v}_\theta(T_j[s])$ is the vector of predicted values for each state $s_t$ in $T_j$, such that:
    \[
    \mathbf{v}_\theta(T_j[s])[t] = v_\theta(s_t)
    \]
    \item $\mathbf{y}_n(T_j)$ is the vector of $n$-step value targets for the same states, computed as:
    \[
    \mathbf{y}_n(T_j)[t] = \sum_{i=0}^{n-1} \gamma^i r_{t+i} + \gamma^n \cdot v_\theta(s_{t+n}) \cdot (1 - \text{term}(T_j, t+n))
    \]
    where $\text{term}(T_j, t+n)$ is an indicator function for whether the episode terminates before $t+n$. In \autoref{tab:train-hyp-tab}, the \texttt{n\_steps} parameter specifies the $n$ used in the bootstrapped $n$-step value loss, while \texttt{disc\_factor} corresponds to the discount factor $\gamma$.

\end{itemize}
\noindent  

The policy loss $L_P$ is the average cross-entropy between the policy distributions predicted by the neural network and the target distributions collected via AZ planning:
\[
L_P(B) = \frac{1}{\sum_{j=1}^m l_j} \sum_{j=1}^m \sum_{t=0}^{l_j - 1} \sum_{a \in \mathcal{A}} -\pi_t(a) \log \pi_\theta(a|s_t)
\]
where $s_t = T_j[s][t]$, $\pi_t = T_j[\pi][t]$ and $l_j$ is the length of the episode stored in $T_j$.

\medskip
\noindent

\subsection{State Embeddings}
The grid-world environments described in \autoref{subsec: setup} were built as custom adaptations of the FrozenLake Gym environment, ensuring compatibility with the standard Gym/Gymnasium API \cite{gymopenai} \cite{gymnasium2023}. 
In Gym, observations are given as a discrete index $d$, which can be mapped to 2D coordinates as:
\begin{equation*}
(x, y) = \left(\left\lfloor \frac{d}{\text{ncols}} \right\rfloor, d \bmod \text{ncols}\right)
\end{equation*}
where ncols denotes the grid width (equal to the height in our square configurations). This 2D coordinate vector is used as input to the neural network.

\begin{figure}[t!]
\centering
\includegraphics[width=\linewidth]{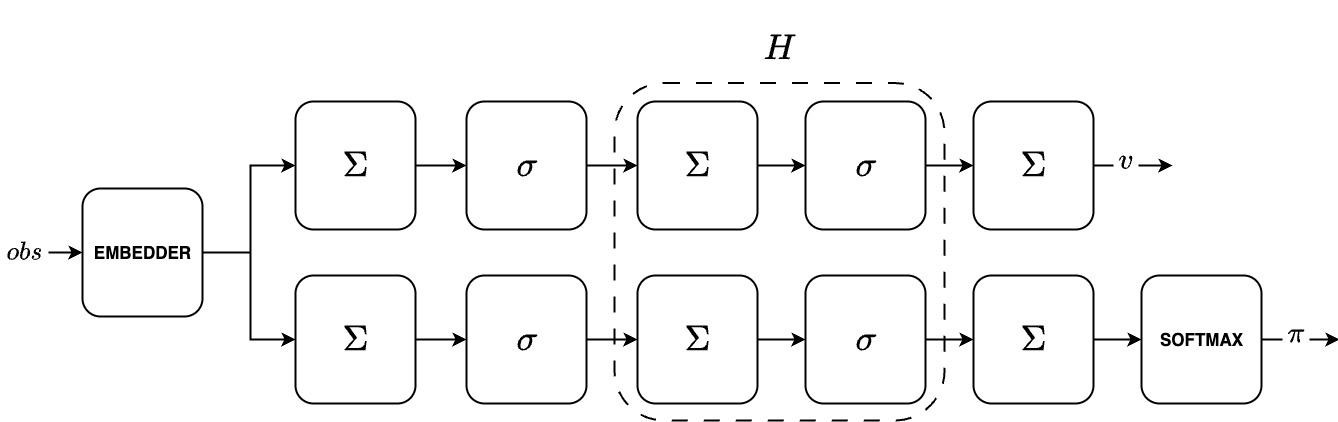}
\caption{MLP architecture with a variable number of hidden layers $H$. The $\Sigma$ block represents a sum and the $\sigma$ block an activation function (e.g., ReLU). The network outputs a value $v$ through the value head and a probability distribution $\pi$ through the policy head.}
\label{fig:nn}
\end{figure}
\subsection{Neural Network Architecture}

Since the input state representations are simple vectors in our GridWorld environment, we adopt a standard Multi-Layer Perceptron (MLP) architecture \cite{mlpandbackprop}. The network features two output heads: one for the value and one for the policy, with the latter passed through a softmax. A visualization of the architecture is shown in \autoref{fig:nn}.

The number of hidden layers $H$ corresponds to the \texttt{hidden\_num} parameter, while \texttt{hidden\_size} defines the neurons per layer. The \texttt{activation} function is always ReLU \cite{relu} in our experiments. 

\subsection{Planning Parameters}
The \texttt{planning\_budget} defines how many node expansions AZ can perform during planning before selecting an action in the real environment, while \texttt{c} corresponds to the selected $C$ exploration parameter of the selection policy. 

To further increase exploration  during training, Dirichlet noise $\eta \sim \text{Dir}(\alpha)$ is added to the prior policy at the root node $x_0$ as: 
\begin{equation*}
\pi^\eta_\theta(a|x_0) = (1 - \epsilon) \pi_\theta(a|x_0) + \epsilon \, \eta,
\end{equation*}
where $\epsilon$ is a mixing parameter. The employed values for these parameters are reported in \autoref{tab:train-hyp-tab} as \texttt{dir\_alpha} and \texttt{dir\_eps}.

\end{document}